%% file: main.tex
\pdfoutput=1
\documentclass{article}

    \PassOptionsToPackage{numbers, compress}{natbib}

\usepackage[final]{neurips_data_2024}

\usepackage[utf8]{inputenc} 
\usepackage[T1]{fontenc}    
\usepackage{hyperref}       
\usepackage{url}            
\usepackage{booktabs}       
\usepackage{amsfonts}       
\usepackage{nicefrac}       
\usepackage{microtype}      
\usepackage{amsmath} 

\usepackage{multirow}
\usepackage[table,xcdraw]{xcolor}
\usepackage[normalem]{ulem}
\useunder{\uline}{\ul}{}

\usepackage{graphicx}
\usepackage{wrapfig}
\usepackage{tikz}
\usepackage{enumitem}
\usepackage{array}
\usepackage{multirow}
\usepackage{colortbl}
\usepackage{booktabs}
\usepackage{pythonhighlight}
\usepackage{listings}
\usepackage{tcolorbox}

\usepackage{verbatim}
\usepackage{listings}
\usepackage{xcolor}

\usepackage{wrapfig}
\usepackage{tikz}
\usepackage{enumitem}
\usepackage{array}
\usepackage{multirow}
\usepackage{colortbl}
\usepackage{booktabs}
\usepackage{pythonhighlight}
\usepackage{listings}
\usepackage{tcolorbox}
\usepackage{graphicx}

\lstdefinelanguage{json}{
  basicstyle=\ttfamily,
  numbers=left,
  numberstyle=\scriptsize,
  stepnumber=1,
  numbersep=8pt,
  showstringspaces=false,
  breaklines=true,
  frame=single,
  backgroundcolor=\color{white},
  literate=
    *{0}{{{\color{blue}0}}}{1}
     {1}{{{\color{blue}1}}}{1}
     {2}{{{\color{blue}2}}}{1}
     {3}{{{\color{blue}3}}}{1}
     {4}{{{\color{blue}4}}}{1}
     {5}{{{\color{blue}5}}}{1}
     {6}{{{\color{blue}6}}}{1}
     {7}{{{\color{blue}7}}}{1}
     {8}{{{\color{blue}8}}}{1}
     {9}{{{\color{blue}9}}}{1}
     {:}{{{\color{red}:}}}{1}
     {,}{{{\color{red},}}}{1}
     {\{}{{{\color{violet}{\{}}}}{1}
     {\}}{{{\color{violet}{\}}}}}{1}
     {[}{{{\color{violet}{[}}}}{1}
     {]}{{{\color{violet}{]}}}}{1},
}
\usepackage{graphicx}
\usepackage{tabularx}
\usepackage{arydshln}
\usepackage{tabularray}
\usepackage{tcolorbox}
\usepackage{geometry} 
\usepackage{adjustbox}
\usepackage{enumitem}

\definecolor{codegreen}{rgb}{0,0.6,0}
\definecolor{codegray}{rgb}{0.5,0.5,0.5}
\definecolor{codepurple}{rgb}{0.58,0,0.82}
\definecolor{backcolour}{rgb}{0.95,0.95,0.92}

\lstdefinestyle{mystyle}{
    backgroundcolor=\color{backcolour},  
    commentstyle=\color{codegreen},
    keywordstyle=\color{magenta},
    stringstyle=\color{codepurple},
    basicstyle=\ttfamily\footnotesize,
    breakatwhitespace=false,         
    breaklines=true,                 
    captionpos=b,                    
    keepspaces=true,                 
    showspaces=false,                
    showstringspaces=false,
    showtabs=false,                  
    tabsize=2
}

\definecolor{keycolor}{rgb}{0,0,0.8}    
\definecolor{stringcolor}{rgb}{0.5,0,0} 
\definecolor{numbercolor}{rgb}{0.5,0,0.5} 

\definecolor{verylightgray}{rgb}{0.9,0.9,0.9}

\lstdefinelanguage{json}{
    basicstyle=\normalfont\ttfamily,
    showstringspaces=false,
    breaklines=true,
    frame=lines,
    backgroundcolor=\color{verylightgray},
    literate=
     *{0}{{{\color{numbercolor}0}}}{1}
      {1}{{{\color{numbercolor}1}}}{1}
      {2}{{{\color{numbercolor}2}}}{1}
      {3}{{{\color{numbercolor}3}}}{1}
      {4}{{{\color{numbercolor}4}}}{1}
      {5}{{{\color{numbercolor}5}}}{1}
      {6}{{{\color{numbercolor}6}}}{1}
      {7}{{{\color{numbercolor}7}}}{1}
      {8}{{{\color{numbercolor}8}}}{1}
      {9}{{{\color{numbercolor}9}}}{1}
      {:}{{{\color{keycolor}{:}}}}{1}
      {,}{{{\color{keycolor}{,}}}}{1}
      {\{}{{{\color{keycolor}{\{}}}}{1}
      {\}}{{{\color{keycolor}{\}}}}}{1}
      {[}{{{\color{keycolor}{[}}}}{1}
      {]}{{{\color{keycolor}{]}}}}{1}
      {"}{{{\color{stringcolor}{"}}}}{1},
}

\definecolor{lightgray}{rgb}{0.94,0.95,0.95}
\lstdefinelanguage{json2}{
    basicstyle=\normalfont\fontfamily{pcr}\selectfont,
    showstringspaces=false,
    breaklines=true,
    frame=lines,
    backgroundcolor=\color{lightgray},
    literate=
     *{0}{{{\color{numbercolor}0}}}{1}
      {1}{{{\color{numbercolor}1}}}{1}
      {2}{{{\color{numbercolor}2}}}{1}
      {3}{{{\color{numbercolor}3}}}{1}
      {4}{{{\color{numbercolor}4}}}{1}
      {5}{{{\color{numbercolor}5}}}{1}
      {6}{{{\color{numbercolor}6}}}{1}
      {7}{{{\color{numbercolor}7}}}{1}
      {8}{{{\color{numbercolor}8}}}{1}
      {9}{{{\color{numbercolor}9}}}{1}
      {:}{{{\color{keycolor}{:}}}}{1}
      {,}{{{\color{keycolor}{,}}}}{1}
      {\{}{{{\color{keycolor}{\{}}}}{1}
      {\}}{{{\color{keycolor}{\}}}}}{1}
      {[}{{{\color{keycolor}{[}}}}{1}
      {]}{{{\color{keycolor}{]}}}}{1}
      {"}{{{\color{stringcolor}{"}}}}{1},
}

\definecolor{mygray}{rgb}{0.95, 0.95, 0.95}
\definecolor{myblue}{rgb}{0.41, 0.50, 0.57}
\definecolor{greyblue}{RGB}{177,221,240}
\definecolor{lightgold}{RGB}{249,247,237}
\definecolor{peacockblue}{RGB}{27,161,226}

\tcbuselibrary{listings,skins,breakable}

\tcbset{
    enhanced,
    breakable,
    colback=mygray, 
    colframe=myblue, 
    boxrule=0.2mm, 
    arc=1mm, 
    top=10pt,
    bottom=10pt,
    left=10pt,
    right=10pt,
    listing only,
    listing options={
        basicstyle=\ttfamily\small, 
        language=Java, 
    },
}

\definecolor{light-gray}{gray}{0.95}

\title{ActionStudio: A Lightweight Framework for Data and Training of Large Action Models}

\author{
\centerline{Jianguo Zhang\thanks{\, Co-first Authors.},~\textbf{Thai Hoang}\footnotemark[1],~Ming Zhu\footnotemark[1],~\textbf{Zuxin Liu}\thanks{\, Core Contributor.},}\\[1mm] 
\centerline{\textbf{Shiyu Wang},~\textbf{Tulika Awalgaonkar},~\textbf{Akshara Prabhakar},~\textbf{Haolin Chen},~\textbf{Weiran Yao},~\textbf{Zhiwei Liu},}\\[1mm] 
\centerline{\textbf{Juntao Tan},~\textbf{Juan Carlos Niebles},~\textbf{Shelby Heinecke},~\textbf{Huan Wang},~\textbf{Silvio Savarese},~\textbf{Caiming Xiong}}\\[3mm]
\centerline{Salesforce AI Research}\\
}

\begin{document}

\maketitle

\begin{abstract}



Large Action models are essential for enabling autonomous agents to perform complex tasks. However, training such models remains challenging due to the diversity of agent environments and the complexity of noisy agentic data. Existing infrastructure offers limited support for scalable, agent-specific fine-tuning and standardized agent data processing. We introduce \textbf{ActionStudio}, a lightweight and extensible data and training framework designed for large action models. ActionStudio unifies diverse agent trajectories using our proposed Unified Format 2.0, supports a range of training workflows with optimized multi-node distributed setup, and integrates robust preprocessing and real-time verification tools. ActionStudio demonstrates up to 9$\times$ higher throughput compared to existing agentic training frameworks, and our trained models yield top performances across public and realistic agent benchmarks. To support the broader research community, we open-source the ActionStudio framework and release \textit{actionstudio-98k}, a curated dataset of 98k high-quality trajectories.\footnote{Code: \url{https://github.com/SalesforceAIResearch/xLAM}}

\end{abstract}

\input{1-introduction}
\input{2-related_work}
\input{3-framework}
\input{4-experiments}
\input{5-conclusion}

\input{6-limitations}

\bibliographystyle{unsrt}
\bibliography{main}

\clearpage
\appendix

\label{sec:appendix}

\input{appendix-2}

\end{document}

%% file: 1-introduction.tex
\section{Introduction}

Action models are becoming increasingly critical for enabling autonomous agents to operate effectively across complex, multi-step tasks in diverse environments-from personal productivity assistants to real-world industrial automation systems. 
While recent open-source initiatives have advanced the action models development~\citep{zeng2023agenttuning,xu2023lemur,zhang2024agentohana,ma2024taco,zhang2024xlam,liu2024agentlite}, infrastructure for efficient agentic data processing and model training remains underdeveloped.

A central challenge lies in the nature of agentic training data, which often comprises long-horizon trajectories with tool interactions, observations, and user feedback originating from varied environments. While prior work has attempted to address data standardization~\citep{zhang2024agentohana,zhang2024xlam}, existing solutions usually rely on instruction-following templates that abstract away tool use~\citep{yin2023lumos,xi2024agentgym} or adopt fixed rigid formats that may not generalize across tasks. Moreover, the data conversion and quality control processes required to turn raw agent trajectories into training-ready datasets are seldom open-sourced, which limits reproducibility and cross-task transfer.

On the training side, general-purpose frameworks such as Transformers~\citep{wolf2020transformers} and LLAMA-Factory~\citep{zheng2024llamafactory} 
have played an important role in Large language model (LLM) development. However, these frameworks 
are primarily designed and optimized for standard LLM workflows, requiring substantial customization and heavy modifications to support agent-specific data and training.
Even high-performing open-source models~\citep{xu2023lemur,zhang2024xlam} have not fully released their implementation code. This creates barriers for researchers and developers aiming effectively to build agentic systems in real-world settings.

To address these challenges, we introduce \textbf{ActionStudio}, an end-to-end, open-source framework for data processing and training of large action models. Designed for production-scale use, ActionStudio integrates a novel \textit{critique-and-filter} pipeline, deterministic rule-based checks, and a real-time verifier for filtering and visualizing agent trajectories. The resulting dataset, \textsc{actionstudio-98k}, comprises 98,000 high-quality trajectories formatted using our proposed Unified Format 2.0. The framework features an extensible backend supporting supervised fine-tuning (SFT) and preference-based learning (e.g., DPO), with flexible configurations including LoRA, quantization, and near-linear multi-node scalability. Models trained with ActionStudio achieve new state-of-the-art results on NexusRaven and the CRM Agent Benchmark, outperforming both open-source agent models and commercial systems.



The key contributions of our work are:
\begin{itemize}[leftmargin=*]
    \item  We present an open-source, lightweight, and efficient agentic training framework, supporting flexible workflows such as LoRA, full fine-tuning, and multi-node distributed training. Our framework achieves up to 9$\times$ higher throughput compared to popular agentic training frameworks.
    
    \item We propose a \textit{critique-and-filter} pipeline and a real-time data verifier that automate the ingestion, filtering, and conversion of diverse agent trajectories. We also release \emph{actionstudio-98k}, a high-quality dataset of 97,755 trajectories spanning over 30,000 APIs and 300 domains, structured using our designed Unified Format 2.0 to facilitate agent research.
    
    \item We demonstrate ActionStudio's effectiveness across public and realistic industry agent benchmarks, showing its utility and practical value for real-world agent applications. 
\end{itemize}
 
\begin{figure*}
  \begin{center}
\includegraphics[width=0.95\linewidth]{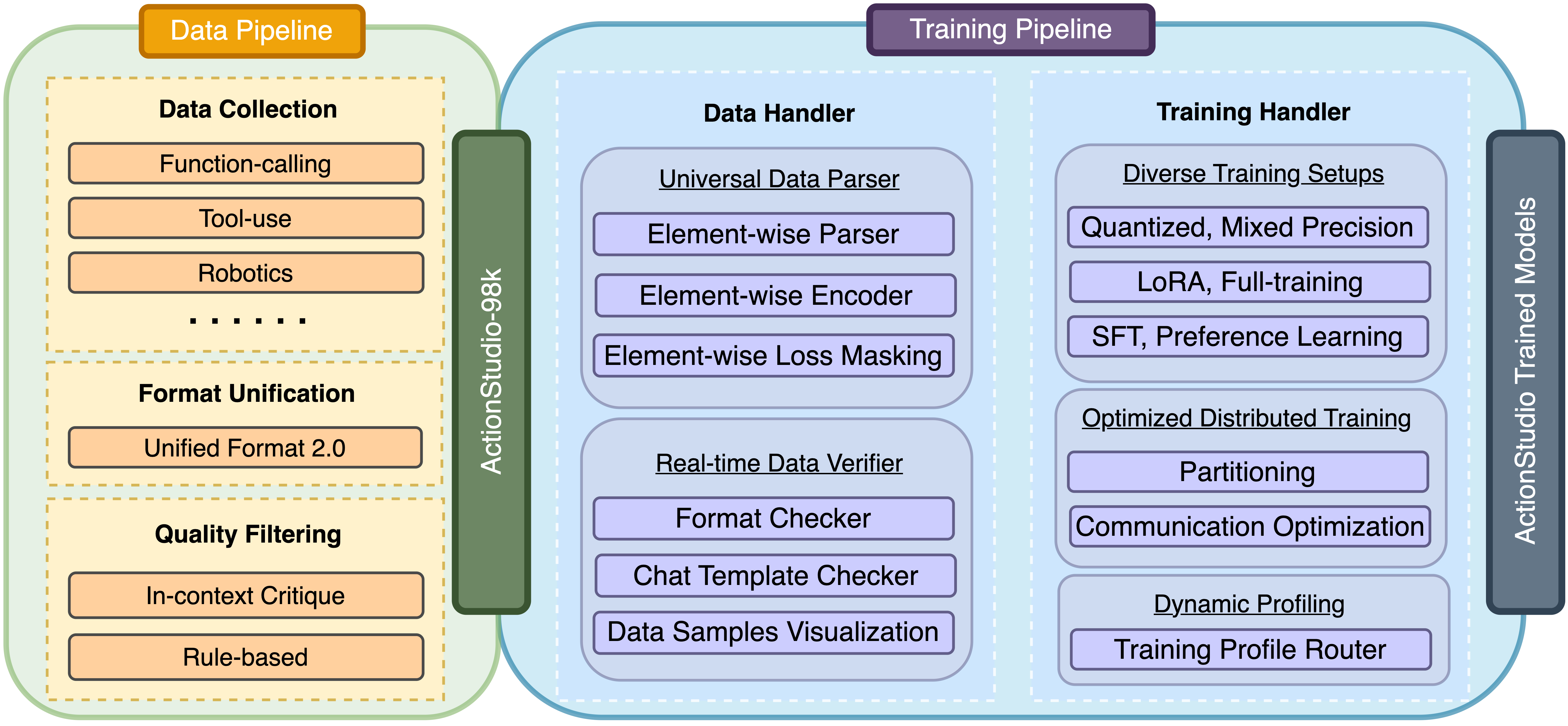}
  \end{center}
    \caption{Framework of ActionStudio.}
  \label{fig:main-framework}
\end{figure*}

%% file: 2-related_work.tex
\section{Related Work}

\subsection{Agent Data}

While proprietary models often restrict data accessibility, the research community has significantly advanced open-source initiatives by releasing extensive agentic datasets~\citep{zeng2023agenttuning,liu2023agentbench,li2023api,zhang2023dialogstudio,xi2024agentgym,song2024agentbank,liu2024toolace,ma2024taco}. However, due to the inherent complexity and heterogeneous nature of agent trajectories across different environments, datasets often vary widely in format, creating substantial hurdles for industrial-scale model development. Recent efforts such as Lumos~\citep{yin2023lumos}, AgentOhana~\citep{zhang2024agentohana}, and xLAM~\citep{zhang2024xlam} have aimed to standardize datasets into unified formats to reduce errors and simplify training. Nonetheless, these initiatives have not fully open-sourced the data conversion and automation pipelines, limiting widespread adoption and scalability.

\subsection{Large Action Models}

Beyond proprietary model APIs, significant progress has been made toward developing open-source large action models, specifically tailored for complex agent-oriented tasks~\citep{xu2023lemur,qin2023toolllm,liu2023bolaa,zhang2024agentohana,ma2024taco,zhang2024xlam,liu2024apigen,liu2024pract,hoang2025lam}. These models have achieved impressive performance on benchmarks, showing the growing capability of open-source initiatives. While established frameworks like Transformers~\citep{wolf2020transformers} and LLAMA-Factory~\citep{zheng2024llamafactory} facilitate general language model training, specialized frameworks designed explicitly for fine-tuning LAMs remain limited. Our work contributes directly to addressing this gap by providing an efficient, lightweight training pipeline within ActionStudio, significantly reducing the complexity and resource requirements associated with training high-performing LAMs.



%% file: 3-framework.tex
\section{Framework}

To support the development of high-performing large action models (LAMs), we present \textbf{ActionStudio}, a comprehensive framework for agentic data constructing and model training under a single, modular system. ActionStudio is composed of two core pipelines: a \emph{data pipeline} for standardizing and preparing diverse agentic data sources, and a \emph{training pipeline} for fine-tuning large language models on agent tasks at various scales. 


\subsection{Data Pipeline} \label{data-pipeline}

The data pipeline in ActionStudio is designed to process diverse agentic data sources into standardized, training-ready formats. It includes four major components: data collection, format unification, quality filtering, and format conversion. This modular structure ensures the framework is extensible, scalable and compatible with a wide range of agent environments and models.

\subsubsection{Data Collection} To support agentic model training, we compile a diverse set of high-quality datasets from multiple agent environments and domains, including but not limited to function calling, tool-use, and robotic agent trajectories. The datasets vary in structure and components, which poses significant challenges for LAM training.

\subsubsection{Format Unification 2.0}
Previous work~\citep{zhang2024agentohana, zhang2024xlam} proposed a modular schema (Unified Format 1.0) to standardize agentic interaction data, with fields such as \texttt{task\_instruction}, \texttt{query}, \texttt{tools}, and a list of \texttt{steps} capturing tool calls, intermediate thoughts, user feedback, and observations. While partially effective for general-purpose processing and augmentation, Unified Format 1.0 was not natively compatible with message-based interfaces expected by most open-source LLMs, resulting in non-trivial conversion and error fixing overhead for training and deployment. Examples of Unified Format 1.0 and its corresponding training prompt are shown in Figures~\ref{fig:unified_format} and~\ref{fig:training_format} in the Appendix.

To address these limitations, we introduce \textbf{Unified Format 2.0}, a redesigned schema that natively aligns with modern chat-based LLM APIs and HuggingFace-style chat templates. Unlike prior work, Unified Format 2.0 is designed to support both \textit{training, inference and evaluation workflows}-including Alpaca-style~\citep{taori2023alpaca} \texttt{(input, output)} pairs, ShareGPT-style~\citep{zhang2023dialogstudio} multi-turn dialogues, and general chat-based interaction formats. Its structure minimizes data transformation overhead, enabling direct use in common fine-tuning pipelines and runtime LLM interfaces.

Unified Format 2.0 introduces new abstractions that modularize agent trajectories into semantically grounded and model-compatible components, such as task instructions, available tools, and user-agent exchanges (including tool calls and execution results). This simplifies downstream formatting, promotes consistency across various data sources, and enables plug-and-play integration into both training and deployment systems. An example in Unified Format 2.0 and its converted chat format are shown in Figures~\ref{fig:unified_format_2} and~\ref{fig:training_format_2} in the Appendix.


\subsubsection{Data Quality Filtering} \label{sec:in-context-critique}
Prior work~\citep{chen2023alpagasus,zhang2024xlam} has explored leveraging LLMs-like GPT-4-class models-for automatically evaluating and filtering trajectory quality, thereby reducing the reliance on manual annotation. These approaches typically score trajectories along dimensions such as correctness, hallucination, tool-use appropriateness, and overall response quality. However, we observe that off-the-shelf LLM evaluators tend to produce overly confident or median-biased scores and often fail to detect subtle or context-dependent hallucinations, and the issues are also noted in prior studies.

To address these limitations, we design a novel agent trajectory quality filtering method based on \textbf{In-Context Critique Filtering}. We augment the LLM evaluator with a small set of curated exemplars illustrating common failure cases and preferred critique behaviors. This simple yet effective in-context guidance leads to more fine-grained, human-aligned evaluations, particularly for ambiguous or borderline trajectories. In addition, we fine-tune open-source models using agent critique data to reduce reliance on commercial LLMs and further improve performance, making quality filtering more cost-effective and accessible. The critique pipeline is complemented by \textit{rule-based} filtering pipeline that catch systematic errors (e.g., missing function calls, wrong function names or arguments, hallucinated agent actions). 

Together, these components provide a scalable, high-precision pipeline for selecting training trajectories. Human verification in Section~\ref{ab:human-verification} demonstrates the effectiveness of the approach.


\subsubsection{ActionStudio-98k}
To facilitate open-source agent research, we release \textit{actionstudio-98k}, a curated collection of 97,755 high-quality agent trajectories spanning diverse environments and task domains. The dataset includes 69,271 single-turn and 28,484 multi-turn trajectories, with multi-turn examples averaging 9 steps per trajectory. Filtered, critiqued, and corrected through the ActionStudio data pipeline, the trajectories are sourced from public agent datasets such as \cite{liu2023agentbench,zhang2023dialogstudio, zhang2024xlam,liu2024toolace, xi2024agentgym, guo2024stabletoolbench} and cover over 30,000 APIs across more than 300 domains. The dataset includes programmatic tool-use sequences, embodied agent interactions, and both single- and multi-turn tasks, all represented in our unified  format 2.0.

\subsection{Training Pipeline}
Our framework is designed to deliver unparalleled versatility and efficiency in agentic training, ensuring it addresses diverse needs while optimizing scalability and performance. The following outlines how these are achieved.
\subsubsection{Data Handler}

\paragraph{\textbf{Universal Data Parser.}}

Training agentic models often involves working with highly diverse data structures, ranging from single-step responses to multi-step reasoning processes, multi-turn conversations, and various role configurations among users and agents (or groups of agents). To manage this complexity and maximize flexibility, our framework uses element-wise parsing and encoding. Each part of the conversation history is parsed as independently as possible, while still following the chat template. This approach simplifies the process of applying fine-grained loss masking and supports different training objectives. As a result, researchers have full control over these processes, making it easy to fine-tune agentic models for specific tasks. Ultimately, this design accelerates experimentation and speeds up model development.

\paragraph{\textbf{Real-Time Data Verifier.}}

Given the wide range of configurations and processing capabilities possible with our framework, it is critical to have robust mechanisms to validate data integrity throughout the training pipeline. The Real-Time Data Verifier keeps training pipeline under-control by dynamically running three checks: the Format Checker instantly flags data instances with missing fields or wrong structures, the Chat Template Checker ensures every conversation fits the provided Chat template, and the Data Samples Visualization presents ``before-and-after" views of each data entry at every stage - before and after preprocessing, applying conversational templates, and encoding. By offering visibility into the transformations applied at each step, the verifier enables users to validate that their data complies with the expected format. This minimizes the risk of unexpected behaviors during training, ensuring a smoother development process and more reliable model performance.

\subsubsection{Training Handler.}
\paragraph{\textbf{Comprehensive Support for Diverse Training Setups}}
We provide an extensive range of functionalities to accommodate virtually any training pipeline for agentic models. From Supervised Instruction Fine-tuning to Human Preference Alignment, from lightweight approaches like quantized training and LoRA (Low-Rank Adaptation)~\citep{hu2021lora} to full-scale training in mixed precision, our framework has it covered. This flexibility allows researchers and practitioners to seamlessly adapt to varying training requirements and computational resource constraints.

\paragraph{\textbf{Highly Optimized Distributed Training at Scale.}}
Our framework is purpose-built for highly efficient training of large-scale agentic models. To this end, heavy effort has been invested in optimizing performance for distributed settings. We have studied communication patterns in common Transformer architectures~\cite{wolf2020transformers}, focusing on reducing inefficiencies in layer-to-layer interactions and communications between experts under Mixture of Experts settings. Additionally, we have optimized GPU-to-GPU communication within a single node as well as cross-node communication, reducing bottlenecks and enabling seamless scalability to industrial-scale training clusters. Users can also benefit from widely adopted parallelization and sharding strategies, such as those offered by DeepSpeed ~\cite{rasley2020deepspeed}, ensuring compatibility with industry-standard training practices. These enhancements ensure that our framework consistently delivers top-tier performance and efficiency, even in demanding distributed environments.

\paragraph{\textbf{Dynamic Profiling for Model-Specific Optimizations.}}
Recognizing the variety of architectures and model sizes in the agentic training space, we have implemented an initial dynamic profiling system to enhance efficiency and improve ease-of-use. When provided with a model checkpoint, our framework dynamically routes it to an optimized configuration tailored for that architecture and model size under the current resource situation. This automation eliminates the need for labor-intensive tuning, allowing researchers to achieve higher efficiency with less manual effort.

%% file: 4-experiments.tex
\section{Experiments}

\input{acl2025_tables/nexusraven}

\subsection{Model Training}

Utilizing the ActionStudio framework, we conducted supervised fine-tuning 
on selected open-source models including Mistral series~\citep{jiang2023mistral, jiang2024mixtral} and Llama 3 series~\citep{grattafiori2024llama}. 
To showcase ActionStudio's effectiveness across various sizes, we fine-tune from smaller-scale models such as Mistral variants to larger-scale models like Llama-3.1-70b-inst and Mixtral-8x22b-inst. 

We set the sequence length between 8k and 16k, the batch size between 32 and 96 and the learning rate between 2e-6 and 2e-4, employing a cosine learning rate scheduler with 5\% warm-up steps. Smaller models were fine-tuned on single NVIDIA H200 pod, while larger models were fine-tuned on both single and more H200 pods for comparison. Each H200 pod is equipped with 8 NVIDIA H200 GPUs, each having 141GB of memory.


\subsection{Benchmarks}
To demonstrate the effectiveness of ActionStudio, we selected NexusRaven and CRM Agent Bench. 

NexusRaven \cite{srinivasan2023nexusraven} provides a diverse benchmark for function calling, comprising 318 test examples across 65 distinct APIs. The dataset is curated through a structured pipeline that mines function definitions, docstrings, and execution contexts from open-source corpora. LLMs are then prompted to generate natural language queries, Chain-of-Thought (CoT) traces, and hard-negative function candidates to enhance evaluation difficulty. NexusRaven specifically evaluates model performance using precision, recall, and F1-score metrics for both function retrieval and argument inference tasks, offering a comprehensive assessment framework for function calling capabilities.

The CRM Agent Benchmark~\footnote{\url{https://www.salesforceairesearch.com/crm-benchmark}} is a proprietary evaluation developed by Salesforce. It assesses proficiency of AI models across critical and real CRM agent scenarios, focusing on accuracy in agent topic identification, accurate generation of function calls, and creation of contextually appropriate free-text responses. This benchmark emphasizes realistic business use cases by incorporating several hundred real CRM data points with expert assessments, providing valuable insights into the practical utility and reliability of language models for commercial deployment. 
Importantly, for all experiments, we fine-tune models exclusively on public datasets, ensuring that no customer data from any companies is utilized during training.

We set the model temperature to 0 during evaluations to ensure deterministic and replicable results.

\subsection{NexusRaven}
Table \ref{tab:nexusraven} shows the comparative performance of various models evaluated on NexusRaven. ActionStudio-trained models consistently outperform baseline and prominent commercial models. In particular, our fine-tuned ActionStudio-Mixtral-8x22b-inst-exp  model achieves the highest overall F1-score (0.969), reflecting strong precision and recall, significantly surpassing commercial alternatives such as GPT-4 and GPT-4o. It also surpasses recently released large-scale models such as GPT-4.1 and Llama-4, both of which highlight strong agentic capabilities.   Similarly, other fine-tuned models also exhibit robust performance. These results show the effectiveness of ActionStudio's pipeline in enhancing function-calling capabilities. 

\subsection{CRM Agent Bench}

\input{acl2025_tables/crm_agent_bench}

Table \ref{table:crm-agent} illustrates our performance on the realistic industry Agent Benchmark. Our ActionStudio-Llama-3.3-70b-inst-exp model achieves the highest overall performance with an average accuracy of 0.87, surpassing the base Llama-3.3-70b-inst model (0.84). This reflects balanced capabilities across all three dimensions. 
Similarly, the ActionStudio-Llama-3.1-70b-inst-exp shows robust performance.

Furthermore, our fine-tuned ActionStudio-Mixtral-8x22b-inst-exp significantly outperforms its base, Mixtral-8x22b-inst, by 11\%.  Models such as ActionStudio-Mistral-7b-inst-exp also exhibit marked improvements of 13\% compared to their instruct baselines, confirming that the pipeline benefits both large and small checkpoints.
Additionally, our models are also ahead of strong agentic models such as o1-preview (0.85) and AgentOhana-8x22b-inst (0.80). These findings show ActionStudio's capability to train versatile and reliable models  for practical and realistic agent scenarios.

\input{acl2025_tables/training_eff}

\subsection{Training Efficiency}
Table \ref{tab:training_efficiency_compact} benchmarks raw training throughput (tokens / s) for three model sizes,  Llama-3.1-8b, Mixtral-8x7b, and Mixtral-8x22b, under the four most commonly used fine-tuning regimes: (i) NF4-quantized LoRA (Q+LoRA), (ii) BF16 LoRA, (iii) full BF16 fine-tuning (FT) on a single pod (FT-1), and (iv) full FT on multiple pods (FT-2/FT-4). For the LoRA setting, we update the q\_proj, k\_proj, v\_proj, and o\_proj layers, with lora\_r set to 32 and lora\_a to 64.  To contextualize these results, we replicated all configurations on the same cluster and GPU infrastructure for two competing systems that support agentic model trainings, \textsc{AgentOhana} ~\cite{zhang2024agentohana} and \textsc{Lumos} ~\cite{yin2023lumos}.
The results from the comparison highlight the efficiency and capability of our framework. 
\paragraph{Quantised LoRA.}
First, we can look at the throughput for quantized-based trainings (Q+LoRA). For the Llama-3.1-8B model, ActionStudio achieves a throughput of 79k tokens/s under Q+LoRA, outperforming AgentOhana (27.9k; 2.8x slower) and Lumos (8.4k; 9.4x slower).
On the medium-sized Mixtral-8x7b, ActionStudio reaches 46k tokens/s, outpacing AgentOhana (25k; 1.9x slower) and Lumos (5k; 9.0x slower).
For the larger configuration, Mixtral-8x22b, ActionStudio sustains 14.7k tokens/s, 1.8x and 8.9x faster than AgentOhana and Lumos, respectively.

\paragraph{BF16 LoRA.}
Under BF16-LoRA setting, ActionStudio also demonstrates a clear advantage. For Llama-3.1-8B, we obtain 76.8k tokens/s compared to AgentOhana (53.7k; 1.4x slower) and Lumos (59.6k; 1.3x slower). For Mixtral-8x7b, our system reaches 47.4k tokens/s, while AgentOhana encounters out-of-memory (OOM) errors; Lumos manages 33.6k tokens/s (1.4x slower). For Mixtral-8x22b, ActionStudio delivers a throughput of 14.7k tokens/s, while AgentOhana fails to complete training under this regime due to OOM and Lumos gets 5.1k tokens/s, which is 2.9x slower than us.

\paragraph{Full-model tuning.}
Next, for full model training, where much more parameters needed to be updated, and a more efficient usage of resources needed to be done in order to enable this. As we can see from the table, ActionStudio is the only framework able to fully update all model parameters across all three model sizes, with full fine-tuning on a single pod is only 16-30\% slower than LoRA. For smaller models like Llama-3.1-8b and Mixtral-8x7b, AgentOhana and Lumos can also support training, but at noticeable slower performance than ActionStudio.

\paragraph{Multi-pod scalability.}
Finally, both AgentOhana and Lumos do not support multi-pods trainings, while in ActionStudio, the throughput remains to be linearly scaled for both 2 and 4 pods: Llama-8B throughput rises from 64 k (1 pod) to 125 k (2 pods) and 224 k (4 pods); similar scaling appears for both Mixtral variants. This not only demonstrate our capability to support larger model trainings with ActionStudio, but also at a very efficient speed.

\paragraph{Long-context support.}
We stress-tested ActionStudio on an extreme setting, Mixtral-8x22b with \texttt{BS/GPU=1, Seq=32k}, and observed \emph{no} throughput loss. In fact, throughput \emph{increased} slightly compared with the standard \texttt{BS/GPU=8, Seq=4k}:
15.2 k vs.\ 14.7 k tokens/s for Q+LoRA (+4 \%),
15.4 k vs.\ 14.7 k for BF16-LoRA (+5 \%),
and 46.9 k vs.\ 44.4 k for FT-4 (+5 \%).
These results confirm that ActionStudio's memory scheduler scales seamlessly to much long token contexts, enabling efficient long-horizon agent training without manual tuning.

In summarization, across the three model sizes and four tuning regimes, ActionStudio is consistently the fastest-up to 9x quicker than \textsc{AgentOhana} and \textsc{Lumos}-and the \emph{only} framework that (i) completes every LoRA configuration without out-of-memory (OOM) failures, (ii) supports full-model tuning on multiple pods with linear speed scaling, and (iii) in our tests under ActionStudio framework, it remains similar throughput when Mixtral-8x22b is trained with a 32k-token context.

\subsection{Ablation Study}
\input{acl2025_tables/ab_testing}


\paragraph{Human Verifications.}\label{ab:human-verification}

To quantify this ActionStudio's data filtering effect, we commissioned an independent third-party annotation provider to audit a uniformly random sample of 150 trajectories, stratified across datasets.  Annotators followed the four rubric items defined in \S\ref{sec:in-context-critique}-\emph{correctness}, \emph{hallucination}, \emph{tool-use appropriateness}, and \emph{overall response quality}-and then indicated whether ActionStudio's keep/remove decision was correct. Agreement between ActionStudio and human judges reached \textbf{85\%}, roughly  15\% over previous LLM-based agent data critiquing baseline. This shows the effectivenes of ActionStudio on handling complex trajectories. Besides, through detailed analysis, the residual 15\% disagreement highlights future scopes such as integrating multi-turn self-consistency checks, domain-specific hallucination detectors, or adaptive thresholds that evolve with new agent behaviors-to push reliability even closer to human parity. 

\paragraph{Ablation on Data Pipelines.}
Table \ref{tab:crm_ablation} examines how data quality influences full-tuning (\textsc{FT}) of the Mixtral-8x22b-inst backbone. We compare four settings that differ only in the data processing pipeline during FT: the untuned baseline, FT on \emph{ActionStudio-processed} trajectories, FT on \emph{raw} agent data trajectories, and FT on \emph{AgentOhana-processed} trajectories.

\textbf{ActionStudio preprocessing yields the largest gains.}
Applying ActionStudio's critique-and-filter pipeline lifts \emph{function-call} accuracy from 0.65 to \textbf{0.75} (+10), \emph{free-text} accuracy from 0.60 to \textbf{0.82} (+22), and the overall score from 0.74 to \textbf{0.85}.

\textbf{Raw data degrades performance.}
Naively fine-tuning on unfiltered logs drives function-call accuracy down to 0.44 and reduces the aggregate metric to 0.72-\emph{worse} than  baseline-illustrating noisy agent trajectories can overwhelm the learning signal.

\textbf{AgentOhana preprocessing is helpful but less effective.}
Cleaning the same corpus with AgentOhana's agent data pipeline partially recovers performance (0.66 / 0.84 / 0.80) yet still lags behind ActionStudio on every metric, implying that ActionStudio data pipeline could better target the error modes of complicated agent trajectories.




%% file: acl2025_tables/nexusraven.tex
\begin{table*}[ht]
    \centering
    \small
    
    \begin{tabular}{l@{\hskip 60pt}c@{\hskip 40pt}c@{\hskip 40pt}c}

        \toprule
        \textbf{Model} & P\textsubscript{api} & R\textsubscript{api} & \textbf{F1\textsubscript{api}} \\
        \midrule    
        ActionStudio-Llama-3.3-70b-inst-exp & 0.950 & 0.953 & \underline{0.951} \\
        ActionStudio-Llama-3.1-70b-inst-exp  & 0.940 & 0.943 & \underline{0.942} \\ 
        \hdashline
        ActionStudio-Mixtral-8x22b-inst-exp  & 0.969 & 0.969 & \textbf{0.969} \\
        ActionStudio-Mistral-latest-12b-inst-exp & 0.953 & 0.956 & \underline{0.954} \\
        ActionStudio-Mistral-7b-inst-exp  & 0.884 & 0.884 & 0.884 \\
        \hline
         Llama-3.3-70b-inst~\citep{grattafiori2024llama}  & 0.917 & 0.934 & 0.925 \\
         Mistral-latest-12b-inst~\citep{jiang2024mixtral}  & 0.906 & 0.940 & 0.923 \\
        Llama-3.1-70b-inst~\citep{grattafiori2024llama}  & 0.907 & 0.915 & 0.911 \\
        GPT-4o-2024-11-20~\citep{hurst2024gpt}  & 0.943 & 0.840 & 0.889 \\
         GPT-4.1-2025-04-14~\citep{openaigpt41}  & 0.841 & 0.846 & 0.843 \\
        DeepSeek-r1-671b~\citep{guo2024deepseek}  & 0.837 & 0.840 & 0.838 \\
        Mistral-7b-inst~\citep{jiang2023mistral}  & 0.814 & 0.827 & 0.821 \\
        Llama-4-Maverick-400b-inst~\citep{metallama4}  & 0.796 & 0.796 & 0.796 \\
        Llama-4-Scout-109b-inst~\citep{metallama4}  & 0.787 & 0.789 & 0.788 \\
        Mixtral-8x22b-inst~\citep{jiang2024mixtral}  & 0.758 & 0.786 & 0.772 \\
        GPT-4~\citep{achiam2023gpt} & 0.894 & 0.635 & 0.743 \\
        \bottomrule
    \end{tabular}
    \caption{Performance comparison on \textbf{NexusRaven}. The best-performing result is indicated in \textbf{bold}, while the second and third-best results are marked with \underline{underline}.}
    \label{tab:nexusraven}
\end{table*}

%% file: acl2025_tables/crm_agent_bench.tex
\begin{table*}[ht]
\centering
\resizebox{1.0\linewidth}{!}{
\begin{tabular}{lccc c}
\toprule
    & Topic Acc 
    & Function Call Acc 
    & Free Text Acc
    & \textbf{Average Acc} \\ 
\midrule
ActionStudio-Llama-3.3-70b-inst-exp & 0.98 & 0.79 & 0.83 & \textbf{0.87} \\
ActionStudio-Llama-3.1-70b-inst-exp & 0.96 & 0.77 & 0.86 & \underline{0.86} \\ 
\hdashline
ActionStudio-Mixtral-8x22b-inst-exp & 0.98 & 0.75 & 0.82 & \underline{0.85} \\
ActionStudio-Mistral-latest-12b-inst-exp & 0.98 & 0.64 & 0.78 & 0.80 \\
ActionStudio-Mistral-7b-inst-exp & 0.95 & 0.49 & 0.74 & 0.73 \\		 
\midrule
DeepSeek-r1-671b~\citep{guo2025deepseek} & 0.82 & 0.83 & 0.94 & \underline{0.86} \\
o1-preview~\citep{jaech2024openai} & 0.98 & 0.75 & 0.81 & \underline{0.85} \\
Llama-3.3-70b-inst~\citep{grattafiori2024llama} & 0.99 & 0.72 & 0.80 & 0.84 \\ 
GPT-4-turbo~\citep{achiam2023gpt} & 0.99 & 0.60 & 0.92 & 0.83 \\ 
Llama-3.1-70b-inst~\citep{grattafiori2024llama} & 1.0 & 0.62 & 0.82 & 0.81 \\ 
AgentOhana-8x22b-inst~\citep{zhang2024agentohana} & 0.90 & 0.66 & 0.84 & 0.80 \\ 
Mixtral-8x22b-inst~\citep{jiang2024mixtral} & 0.98 & 0.65 & 0.60 & 0.74 \\ 
GPT-4o-mini~\citep{hurst2024gpt} & 0.94 & 0.42 & 0.81 & 0.72 \\
Mistral-latest-12b-inst~\citep{jiang2024mixtral} & 0.96 & 0.18 & 0.70 & 0.61 \\ 
Mistral-7b-inst~\citep{jiang2023mistral} & 0.99 & 0.19 & 0.63 & 0.60 \\ 
\bottomrule
\end{tabular}
}
\caption{Accuracy on the \textbf{CRM Agent Benchmark}. The best-performing result is indicated in \textbf{bold}, while the second and third-best results are marked with \underline{underline}.}\label{table:crm-agent}
\end{table*}

%% file: acl2025_tables/training_eff.tex
\begin{table*}[ht]
\resizebox{\linewidth}{!}{
\begin{tabular}{lllllll}
\toprule
\textbf{Training Setup} & \textbf{Framework} & Q\,+\,LoRA & LoRA & FT 1 pod & FT 2 pods & FT 4 pods \\ \midrule
\multirow{3}{*}{Llama-3.1 8b (BS/GPU=6, Seq=8k)}
  & \textbf{ActionStudio (Ours)}            & \textbf{79,306}     & \textbf{76,766}     & \textbf{64,179}     & \textbf{125,097}     & \textbf{224,192} \\
  & AgentOhana               & 27,868\ \textcolor{red}{(\,-65\%)}  & 53,718\ \textcolor{red}{(\,-30\%)}  & 38,550\ \textcolor{red}{(\,-40\%)}  & \dashuline{Not Sup.} & \dashuline{Not Sup.} \\
  & Lumos (Ai2)                   & 8,399\ \textcolor{red}{(\,-89\%)}   & 59,578\ \textcolor{red}{(\,-22\%)}  & 52,852\ \textcolor{red}{(\,-18\%)} & \dashuline{Not Sup.} & \dashuline{Not Sup.} \\ \midrule
\multirow{3}{*}{Mixtral-8x7b (BS/GPU=8, Seq=4k)}
  & \textbf{ActionStudio (Ours)}            & \textbf{46,193}     & \textbf{47,404}     & \textbf{33,661}     & \textbf{71,858}     & \textbf{137,146} \\
  & AgentOhana               & 24,966\ \textcolor{red}{(\,-46\%)}  & \dashuline{OOM}    & \dashuline{OOM}     & \dashuline{Not Sup.} & \dashuline{Not Sup.} \\
  & Lumos (Ai2)                      & 5,115\ \textcolor{red}{(\,-89\%)}   & 33,608\ \textcolor{red}{(\,-29\%)} & 8,151\ \textcolor{red}{(\,-76\%)}  & \dashuline{Not Sup.} & \dashuline{Not Sup.} \\ \midrule
\multirow{3}{*}{Mixtral-8x22b (BS/GPU=8, Seq=4k)}
  & \textbf{ActionStudio (Ours)}            & \textbf{14,703}     & \textbf{14,654}     & \dashuline{OOM}       & \dashuline{OOM}        & \textbf{44,438} \\
  & AgentOhana               & 8,375\ \textcolor{red}{(\,-43\%)}   & \dashuline{OOM}    & \dashuline{OOM}                & \dashuline{Not Sup.} & \dashuline{Not Sup.} \\
  & Lumos (Ai2)                     & 1,660\ \textcolor{red}{(\,-89\%)}   & 5,072\ \textcolor{red}{(\,-65\%)}  & \dashuline{OOM}                  & \dashuline{Not Sup.} & \dashuline{Not Sup.} \\ \hline
  Mixtral-8x22b (BS/GPU=1, Seq=32k) &\textbf{ActionStudio (Ours)}&15,236&15,430&\dashuline{OOM}&\dashuline{OOM}& 46,861\\
\bottomrule
\end{tabular}}
\caption{Training throughput (tokens/s) for each setup under our \textbf{ActionStudio}, the \textbf{AgentOhana} and \textbf{Lumos (Ai2)} frameworks. "\dashuline{OOM}" indicates an out-of-memory error. "\dashuline{Not Sup.}" denotes that the feature configuration is unsupported by the current version of the framework.}
\label{tab:training_efficiency_compact}
\end{table*}

%% file: acl2025_tables/ab_testing.tex


\begin{table*}[ht]
\centering
\small
\resizebox{0.9\linewidth}{!}{%
\begin{tabular}{lcccc}
\toprule
\textbf{Model / Data}          & Topic Acc & Function Call Ac & Free Text Acc & \textbf{Average Acc} \\ 
\midrule
Mixtral-8x22b-inst             &\textbf{0.98}& 0.65 & 0.60 & 0.74 \\ 
\hdashline
FT on ActionStudio (processed)  & \textbf{0.98} & \textbf{0.75} & \textbf{0.82} & \textbf{0.85} \\ 
FT on Raw data                  & \textbf{0.98} & 0.44 & 0.75 & 0.72 \\ 
FT on AgentOhana (processed)    & 0.90 & 0.66 & 0.84 & 0.80 \\ 
\bottomrule
\end{tabular}%
}
\caption{Accuracy on the \textbf{CRM Agent Benchmark}. 
\emph{FT} denotes full-tuning; \textbf{ActionStudio} and \textbf{AgentOhana} are two different training frameworks (each with its own data processing pipeline). Best score per column is \textbf{bold}.}
\label{tab:crm_ablation}
\end{table*}

%% file: 5-conclusion.tex
\section{Conclusion}

We introduced ActionStudio, a lightweight and flexible framework for training large action models. By integrating structured data preprocessing, advanced fine-tuning, and distributed training, ActionStudio simplifies agentic model development. Evaluations on NexusRaven and the  CRM Agent Benchmark, which specifically reflects realistic industry agent scenarios, demonstrated its effectiveness and practical value for robust agentic model solutions.


%% file: 6-limitations.tex
\section{Limitations}

While ActionStudio provides a practical and extensible framework for developing robust large action models for complex agent scenarios, a few limitations remain.  The current implementation primarily focuses on text-based and function-calling scenarios, with future support planned for multimodal and embodied environments. Additionally, although the framework includes standardized formats and a robust data processing pipeline, model effectiveness still depends on the quality and diversity of input datasets, which can vary across use cases. Despite these challenges, ActionStudio significantly lowers the barrier to LAM development and offers a scalable foundation for further innovation in industry and research.

%% file: appendix-2.tex
\section{Appendix}

\subsection{Unified Format 2.0}

To support diverse agentic tasks in a model-friendly way, we introduce Unified Format 2.0, an upgraded version of the format used in prior work. While Unified Format 1.0 was designed to modularize agentic trajectories for general-purpose processing, it lacked alignment with the message-passing format expected by modern LLM APIs. Unified Format 2.0 is designed to be natively compatible with modern chat-based LLM APIs and HuggingFace-style chat templates, significantly reducing the effort required to convert raw data into model-ready training samples. An example of a trajectory in Unified Format 2.0 is shown in Figure \ref{fig:unified_format_2}.

Unified Format 1.0 \cite{zhang2024xlam} introduced a modular schema for representing agentic interaction data, including fields such as \texttt{task\_instruction}, \texttt{query}, \texttt{tools}, and a list of \texttt{steps} that capture tool calls, intermediate thoughts, user follow-ups, and observations. While effective for general processing and augmentation, this format was not directly aligned with the message-based structure expected by most open-source LLMs, requiring non-trivial conversion logic to adapt the data for training. An example showcasing Unified Format 1.0 is presented in Figure \ref{fig:unified_format}. 

In contrast, Unified Format 2.0 structure is based on the conversational schema commonly used in APIs like OpenAI and HuggingFace. It replaces the \texttt{steps} field with a \texttt{conversation} list, where each entry is a message with a specific role (e.g., \texttt{system}, \texttt{user}, \texttt{assistant}, or \texttt{tool}). Tool calls are now explicitly represented inside assistant messages using a \texttt{tool\_calls} field, and tool responses are mapped to messages with the role \texttt{tool}, linking back via a \texttt{tool\_call\_id}. This structure is more compatible with LLM APIs and chat templates, which removes the need for custom scripts to flatten or restructure training samples. Figure \ref{fig:training_format_2} demonstrates that with Unified Format 2.0, the training format can be flexibly changed by applying the corresponding chat template from the tokenizer. In contrary, Figure \ref{fig:training_format}  shows the fixed training format from Unified Format 1.0, which remains unchanged across different models, requiring substantial effort in data format conversion for both training phase and deployment phase.  

\subsection{Evaluation Details for NexusRaven}

To evaluate function-calling performance on the NexusRaven benchmark, we follow a two-step process: (1) parsing tool calls from the model's output, and (2) computing precision, recall, and F1 scores by comparing the parsed predictions to ground-truth annotations.

\noindent \textbf{Tool Call Parsing.} NexusRaven includes a custom parser designed to extract structured tool call information from raw model outputs. Since different models may follow varying output formats (e.g., enclosing tool calls in special tags, including JSON blocks with markdown fencing, or appending irrelevant tokens), the parser applies a series of heuristics to sanitize the output and isolate the tool call content. It then parses the cleaned text into a structured format containing: 1) the function/tool name, 2) the arguments as a dictionary of key-value pairs, and 3) an optional tool call ID. The parser handles edge cases such as missing fields, extraneous formatting tags (e.g., \texttt{<think>}, \texttt{<tool\_call>}), and malformed JSON. 

\noindent \textbf{Metric Computation.} After extracting the predicted and ground-truth tool calls, we compute evaluation metrics at the API level level, which includes the Precision (P\_api), Recall (R\_api), and F1 Score (F1\_api). P\_api and R\_api calculate the proportion of predicted tool calls whose function name matches one in the ground-truth, and the proportion of ground-truth tool calls that were successfully predicted, respectively. The F1\_api is the harmonic mean of API precision and recall. This enables consistent and scalable comparison of function-calling capabilities across models, while maintaining tolerance to minor formatting differences.


\input{figures_tables/fig_unified_format_2}

\input{figures_tables/fig_training_format_2}

\input{figures_tables/fig_unified_format}

\input{figures_tables/fig_training_template}

%% file: figures_tables/fig_unified_format_2.tex
\begin{figure*}[ht]
\begin{lstlisting}[language=json,basicstyle=\scriptsize\ttfamily, backgroundcolor=\color{lightgold!50}]
{
    "unique_trajectory_id": "id",
    "task_instruction": "...",
    "tools": [
        {
            "type": "function",
            "function": {
                "name": "get_sleep_stats",
                "description": "Get the user's sleep statistics for a specified time period.",
                "parameters": {
                    "type": "object",
                    "properties": {
                        "user_id": {
                            "type": "string",
                            "description": "Unique identifier of the user whose sleep statistics will be retrieved.",
                        },
                    },
                    "required": [
                        "user_id",
                    ]
                }
            }
        },
    ],
    "conversation": [
        {
            "role": "user",
            "content": "I would like to get my sleep statistics from last night."
        },
        {
            "role": "assistant",
            "content": "",
            "tool_calls": [
                {
                    "type": "function",
                    "function": {
                        "name": "get_sleep_stats",
                        "arguments": {
                            "user_id": "1234",
                        }
                    },
                    "id": "808380806"
                }
            ]
        },
        {
            "role": "tool",
            "name": "get_sleep_stats",
            "content": {
                "data": {
                    "message": "..."
                }
            },
            "tool_call_id": "808380806"
        },
        {
            "role": "assistant",
            "content": "Your sleep statistics from last night has been retrived successfully."
        }
    ]
}
\end{lstlisting}
\caption{Unified format 2.0 for function calling data.}
\label{fig:unified_format_2}
\end{figure*}

%% file: figures_tables/fig_training_format_2.tex
\tcbset{
    mybox/.style={
        colback=lightgold!50,
        colframe=lightgold!50,
        arc=4mm,
        boxrule=0.5mm,
        left=1mm,
        right=1mm,
        top=1mm,
        bottom=1mm,
        boxsep=1mm,
        coltitle=black,
        sharp corners,
    }
}
\begin{figure*}[htbp]

\begin{tcolorbox}[mybox,  fonttitle=\bfseries\footnotesize, before upper=\small]
\textbf{Prompt:}
\begin{verbatim}
<|begin_of_text|><|start_header_id|>system<|end_header_id|>

Environment: ipython
Cutting Knowledge Date: December 2023
Today Date: 26 Jul 2024

<|eot_id|><|start_header_id|>user<|end_header_id|>

Given the following functions, please respond with a JSON for a function call 
with its proper arguments that best answers the given prompt.

Respond in the format
{"name": function name, 
"arguments": dictionary of argument name and its value}. 
Do not use variables.

{
    "type": "function",
    "function": {
        "name": "get_sleep_stats",
        "description": "Get the user's sleep statistics 
        for a specified time period.",
        "parameters": {
            "type": "object",
            "properties": {
                "user_id": {
                    "type": "string",
                    "description": "Unique identifier of the user whose sleep
                    statistics will be retrieved."
                }
            },
            "required": [
                "user_id"
            ]
        }
    }
}

I would like to get my sleep statistics from last night.<|eot_id|>
\end{verbatim}

\textbf{Output:}
\begin{verbatim}
[{"name": "get_sleep_stats", "arguments": {"user_id": "1234"}}]
\end{verbatim}

\end{tcolorbox}
\caption{\small Example prompt and output for function-calling from unified format 2.0, by applying Llama-3.1-70B-Instruct chat template.}
\label{fig:training_format_2}
\end{figure*}

%% file: figures_tables/fig_unified_format.tex
\begin{figure*}[ht]
\begin{lstlisting}[language=json,basicstyle=\scriptsize\ttfamily, backgroundcolor=\color{lightgold!50}]
{
    "unique_trajectory_id": "id",
    "task_instruction": "...",
    "few_shot_examples": [],
    "query": "The task or the question that the user provides.",
    "tools": [
        {
            "name": "api_name1",
            "description": "description of this api",
            "parameters": {
                "param1": {
                    "type": "string",
                    "description": "",
                },
            }
        },
    ],
    "steps": [
        {
            "thought": "thinking and/or planning process",
            "tool_calls": [
                {
                    "name": "api_name1",
                    "arguments": {
                        "argument1": "xxx.",
                        "argument2": "xxx"
                    }
                }
            ],
            "step_id": 1,
            "next_observation": "observations or feedbacks from the environment/APIs after execution function."
            "user_input": "User follow up input at this turn if any."
        },
    ],
}
\end{lstlisting}
\caption{Unified format 1.0 for function calling data.}
\label{fig:unified_format}
\end{figure*}

%% file: figures_tables/fig_training_template.tex
\tcbset{
    mybox/.style={
        colback=lightgold!50,
        colframe=lightgold!50,
        arc=4mm,
        boxrule=0.5mm,
        left=1mm,
        right=1mm,
        top=1mm,
        bottom=1mm,
        boxsep=1mm,
        coltitle=black,
        sharp corners,
    }
}
\begin{figure*}[htbp]

\begin{tcolorbox}[mybox,  fonttitle=\bfseries\footnotesize, before upper=\small]
\textbf{Prompt:}
\begin{verbatim}
[BEGIN OF TASK INSTRUCTION]
Based on the previous context and API request history, generate an API 
request or a response as an AI assistant. 
[END OF TASK INSTRUCTION]

[BEGIN OF AVAILABLE TOOLS]
[
    {
        "name": "get_fire_info",
        "description": "Query the latest wildfire information",
        "parameters": {
            "location": {
                "type": "string",
                "description": "Location of the wildfire.",
                "required": true,
            },
            "radius": {
                "type": "number",
                "description": "The radius (in miles) around the location.",
            }
        },
    },...
]
[END OF AVAILABLE TOOLS]

[BEGIN OF FORMAT INSTRUCTION]
Your output should be in the JSON format, which specifies a list of
function calls. The example format is as follows. Please make sure the 
parameter type is correct. If no function call is needed, please make 
tool_calls an empty list "[]".
{"thought": "the thought process, or an empty string", "tool_calls": 
[{"name": "api_name1", "arguments": {"argument1": "value1", "argument2":
"value2"}}]}
[END OF FORMAT INSTRUCTION]

[BEGIN OF QUERY]
Can you give me the latest information on the wildfires occurring in California?
[END OF QUERY]

[BEGIN OF HISTORY STEPS]
[
    {
        "thought": "Sure, what is the radius (in miles) around the location of 
        the wildfire?",
        "tool_calls": [],
        "step_id": 1,
        "next_observation": "",
        "user_input": "User: Let me think... 50 miles."
    },
]
[END OF HISTORY STEPS]
\end{verbatim}

\textbf{Output:}
\begin{verbatim}
{"thought": "", "tool_calls": [{"name": "get_fire_info", 
"arguments": {"location": "California", "radius": 50}}]}
\end{verbatim}

\end{tcolorbox}
\caption{\small Example prompt and output for function-calling from unified format 1.0.}\label{fig:training_format}
\end{figure*}